\documentclass[review]{elsarticle}
\usepackage{graphicx}
\usepackage{graphicx}
\usepackage{color}
\usepackage{lineno,hyperref}
\usepackage{import}
\usepackage{algorithmic}
\usepackage{algorithm}
\usepackage{amsmath,amssymb,amsfonts}
\usepackage{movie15}

\begin{document}

\begin{frontmatter}

\title{A Computational Model for Machine Thinking}

\author{Slimane Larabi}

\address{RIIMA Laboratory, Computer Science Faculty, \\
USTHB University, 16111, Algeria}

\begin{abstract}

A machine thinking model is proposed in this report based on recent advances of computer vision and the recent results of neuroscience devoted to brain understanding.
We deliver the result of machine thinking in the form of sentences of natural-language or drawn sketches either informative or decisional. This result is obtained from a reasoning performed on new acquired data and memorized data.

\end{abstract}


\end{frontmatter}


\section{Introduction}\label{sec:introduction}

"Giving the faculty of thinking for a machine" is a challenging research area. Indeed, robots endowed by the perception and hearing senses will be autonomous and then able to make themselves the suitable decision.
There are a plenty of domains where machine thinking is necessary and useful for human development.

In this report, we propose a computational model for machine thinking based on recent advances of computer vision and recent results of neuroscience devoted to brain understanding.
Our model delivers result of thinking in the form of sentences of natural-language or drawn sketches either informative or decisional. This result is obtained from a reasoning performed on new acquired data and memorized data.

We are not concerned about the way humans ‘actually’ think, but we are inspired simply from what come from a lifelong experience with the way we think and act \cite{Bonnefon2020}. Consequently, our purpose does not require a state-of-the-art review of dual-process models.

In this paper, we explain our point of view of machine thinking. A synoptic scheme is proposed and all functional units composing the model are explained.

The proposed model is described in section 2. We conclude this paper with some suggestions and future works.

\section{Machine Thinking: The proposed model}

\subsection{The Meaning of Machine Thinking}

We believe that thinking for machines may be done similarly to thinking for brain. Due to the absence of understanding of brain thinking, we have been inspired by recent works of neuroscience and our comprehension of brain thinking.

\begin{figure}[ht]
\centering
\includemovie[
  poster,controls, loop,
 label=fig1
  text={\small(Click here to see Output of Machine Thinking.)}
]{6cm}{6cm}{something-0.mp4}
\includegraphics[width=6 cm]{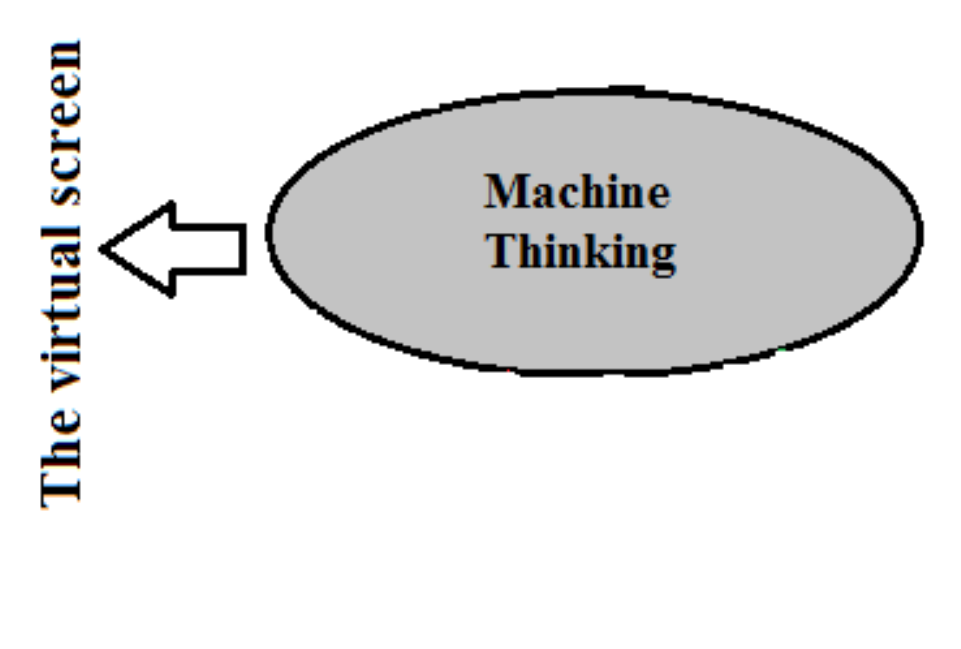}
\caption{Output of Machine Thinking.}
\label{fig1}
\end{figure}

We define \textbf{Machine Thinking } as the inference of decisional and informative information from many sources of data and the deliverance of the result as natural-language sentences and sketches drawing (see figure \ref{fig1}).

\subsection{Elements of the model of Machine Thinking}

\begin{figure}[h!]
	\centering
	\includegraphics[width=8 cm]{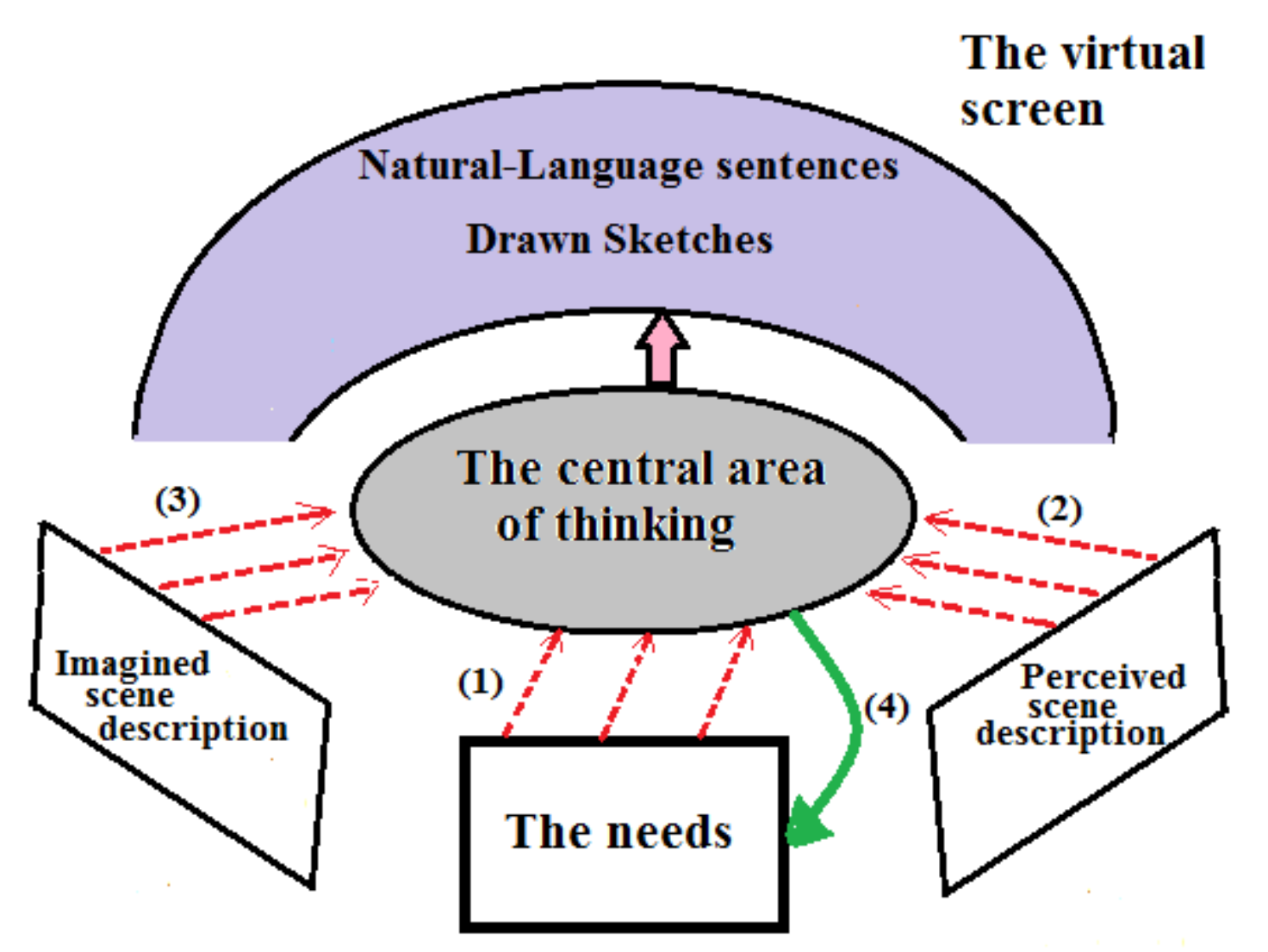}
\caption{Overview diagram of Machine Thinking algorithms. Machine Thinking is a subset of artificial intelligence.}
     \label{fig2}
\end{figure}

The synoptic scheme of the proposed model, illustrated by figure \ref{fig2}, is composed by a Central Area of Thinking (CAT) and three main units.
Machine thinking is performed by the central area of thinking which process information received from:\\
- The unit of "The needs".\\
- The unit of "Description of the perceived scene from images and audio".\\
- The unit of "Description of the imagined scene".\\

Depending on the needs, the scene perceived or in mind, and the knowledge which has been learned, this central area delivers information from many sources and writes it and/or draws it on a virtual screen . Note that the generated information is informative or decisional.

The thinking is started, as illustrated by figure \ref{fig2}, if there are some stimulus sent by:\\
- "The needs" unit, when is active it sends the query to central area of thinking. \\
- "The perceived scene" unit, send continually the description of the perceived scene.\\
- "The imagined scene" unit, send the corresponding description when the machine is in the state of imagination.\\

The central area of thinking uses reasoning and all received information in order to infer a summary of the current state and a set of actions to be performed for satisfying the "needs" unit.

\subsection{The "needs" unit}

\begin{figure}[h!]
	\centering
	\includegraphics[width=6 cm]{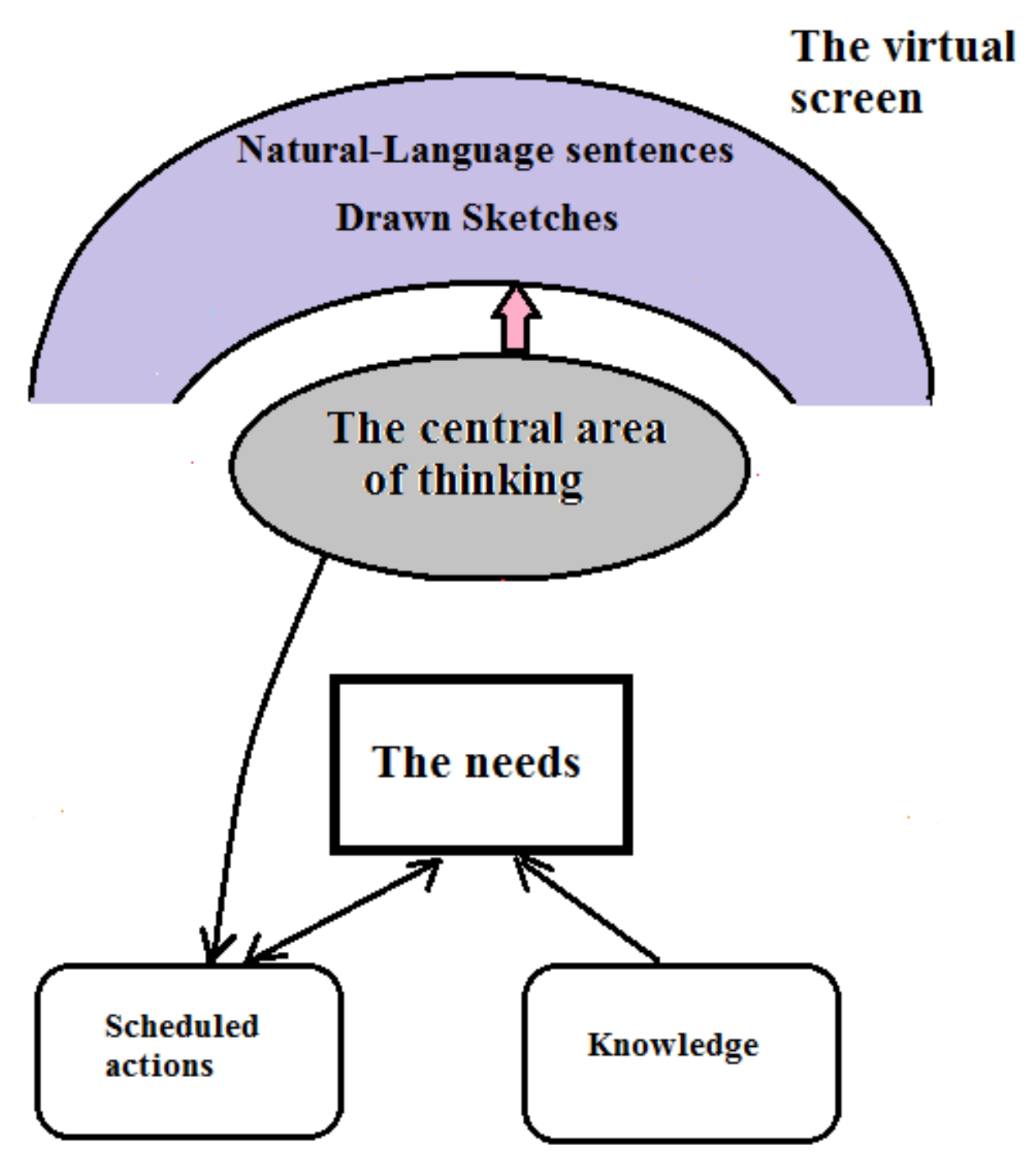}
\caption{Overview diagram of the "Needs" unit.}
     \label{fig3}
\end{figure}

The "needs" is a unit working with a set of data produced or generated during time.
Its known that human expresses needs with external and internal senses\cite{Karjus2021}. Essentially these data are:\\
- The road map of actions to be performed, scheduled after a machine thinking. These actions are transferred from the principal area to this unit and scheduled (see figure \ref{fig3}).\\
- The knowledge acquired by the machine and represented in the "Knowledge" unit.\\

\subsection{The "perceived scene" and the "imagined scene" units}

The "perceived scene" is a unit which processes the acquired images, the heard audio signals and produces a set of natural-language sentences describing the scene. The recent advances in computer vision using neural approaches may be used such as YOLO for object detection and recognition \cite{Redmon2016} \cite{zatout2019ego} and \cite{Aditya2018} \cite{Su2019} \cite{Zhang2021} for image translation to sentences and paragraphs.

The "imagined scene" \cite{Daniel2019} is a kind of scene which provides a tool to "central area of thinking" for reasoning .
Neuroimaging and neuropsychological studies have identified several brain areas that seem to be particularly engaged during the viewing and imagination of scenes including the ventromedial prefrontal cortex, the anterior hippocampus, the posterior parahippocampal cortex, and the retrosplenial cortex \cite{Monk2020}.

Recent studies try to find a computational model for imagined scene activation. The European Research Council and the Human Brain Project, shows how different neurons in the brain work together to store our experiences so that we can later conjure up a mental image and even imagine how a place would appear from a vantage point we haven’t experienced ourselves \cite{Bicanski2018}. They built a computational model that shows how the neuronal activity across multiple brain regions underlying such an experience could be encoded and subsequently used to enable re-imagination of the event.

\subsection{The Central Area of Thinking}

Thinking which is the ability to reach logical conclusions on the basis of prior information is central to human cognition \cite{Goel2017}. Therefore, The advent of neuroimaging techniques has increased the number of studies related the study of examined brain function \cite{Bartley2019} and to the neural basis of deductive reasoning \cite{Wang2020}.
In the proposed model, thinking is performed as a reasoning based on the knowledge.

\section{Conclusion and future works}\label{conclusion}

In this paper we propose a new model for machine thinking. This model is based on three units: The needs, Perceived scene, Imagined scene.
The thinking is a process using all received information and reasoning.
The next steps consist to implement the different modules.

\bibliographystyle{elsarticle-num}

\end{document}